\documentclass{article}
\usepackage{ijcai21}

\usepackage{times}

\usepackage{soul}
\usepackage{url}
\usepackage[hidelinks]{hyperref}
\usepackage[utf8]{inputenc}
\usepackage[small]{caption}
\usepackage{graphicx}
\usepackage{amsmath}
\usepackage{booktabs}
\usepackage{placeins}
\title{Yseop at FinSim-3 Shared Task 2021: Specializing Financial Domain Learning with Phrase Representations}

\author{
Hanna Abi Akl\and
Dominique Mariko\and
Hugues de Mazancourt\\
\affiliations
Yseop\\
\emails
\{habi-akl, dmariko, hdemazancourt\}@yseop.com
}

\begin{document}

\maketitle

\begin{abstract}
In this paper, we present our approaches for the FinSim-3 Shared Task 2021: Learning Semantic Similarities for the Financial Domain. The aim of this shared task is to correctly classify a list of given terms from the financial domain into the most relevant hypernym (or top-level) concept in an external ontology. For our system submission, we evaluate two methods: a Sentence-RoBERTa (SRoBERTa) embeddings model pre-trained on a custom corpus, and a dual word-sentence embeddings model that builds on the first method by improving the proposed baseline word embeddings construction using the FastText model to boost the classification performance. Our system ranks 2$^{nd}$ overall on both metrics, scoring 0.917 on Average Accuracy and 1.141 on Mean Rank.
\end{abstract}

\section{Introduction}

A hypernym or hyperonym is a concept which is superordinate to another one. In computer science, it is often represented as an \textit{IS-A} relationship. For example, \textit{animal} is a hypernym of \textit{cat} and \textit{equity index} is a hypernym of \textit{S\&P 500 Index} \cite{murphy_2003}.
Hypernymy, i.e. the capability to relate generic terms or classes to their specific instances, lies at the core of human cognition \cite{camacho-collados-etal-2018-semeval}. Hypernymy modeling has been widely studied in natural language processing (NLP) for decades. Particularly, results based on embeddings methods \cite{henderson2017learning,nguyen-etal-2017-hierarchical,wang-he-2020-birre,10.5555/2832415.2832443} show promise but the challenge remains in specializing these embeddings in particular areas such as the financial domain because of different aspects of language such as precise terms (e.g. abbreviations) and specific semantics that are badly or not covered at all by general-purpose models.
\par The FinSim 2020 shared task \cite{maarouf-etal-2020-finsim} was the first task that attempts to combine hypernym classification methods in the financial domain. The FinSim-3 Shared Task 2021: Learning Semantic Similarities for the Financial Domain iterates on the previous editions by proposing an extended dataset with more diversified financial concepts.
\par In this paper, we present our approaches which focus on domain-specific learning embeddings using as little data as possible. Although the shared task permits the use of external sources, we limit our training to the Financial Industry Business Ontology (FIBO)\footnote{https://spec.edmcouncil.org/fibo/} data as well as the set of prospectuses in English curated and made available by the organizers. The corpus size for the latter set is estimated to about 10 million tokens. We explore two methods: the first is based on a custom sentence-level embeddings training using SRoBERTa \cite{reimers2019sentencebert} and a term-definition dataset compiled from the FIBO website, and the second is a concatenated sentence-word embeddings model combining the custom SRoBERTa embeddings with a FastText\footnote{https://radimrehurek.com/gensim/models/fasttext.html} word embeddings model trained on the prospectuses set and the constructed FIBO dataset.
\par We also explore and compare empirically the performance of several classifiers. Our experimental results demonstrate that while the domain-specific custom embeddings enhance the classification performance, class imbalances still hinder the recognition of under-represented classes. We analyze these results based on the number of labels provided in the training dataset as well as those extracted from the FIBO website.
\par The rest of this paper is organized as follows. Section 2 introduces the technical details of our proposed approaches. Section 3 empirically evaluates the performances of our methods and presents our results. Section 4 provides the conclusions of our work.

\section{Proposed Approaches}

We make use of custom corpus and exploit sentence-level and word-level embeddings in the context of phrase representation learning. We also test several classifiers in our term classification approaches. The general framework is shown in Figure 1. This framework consists of customized corpus collections, sentence and word representation learning methods and term classification strategies. We will elaborate on each component below.

\FloatBarrier
\begin{figure*}[!ht]
    \centering
    \noindent{\includegraphics[width=\textwidth]{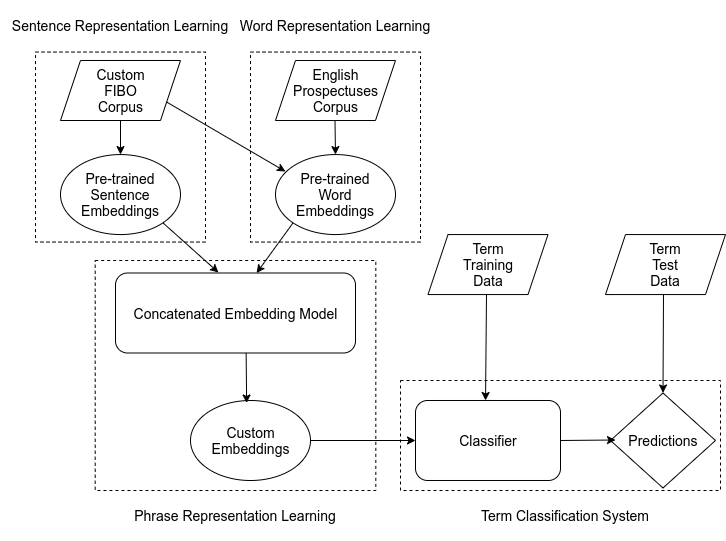}}
    \caption{Framework of our proposed approach}
\end{figure*}
\FloatBarrier

\subsection{Customized Corpus Collection}

General word embeddings are trained on domain-independent corpus. However, different domains have their proper semantics. In order to learn domain-specific representations for financial data, we base our work on collected customized corpus. We use the set of English prospectuses provided by the shared task organizers that contains 203 documents which roughly amounts to an estimated size of 10 million tokens. We augment this set with an extracted corpus from the FIBO website that we also use to train the sentence embeddings.
\par Sentence embeddings already contain contextual information. However, they suffer from the same domain specialization problem as word embeddings. We choose to work with a specialized corpus to generate our sentence embeddings and use the FIBO website provided by the shared task organizers. Starting from the predefined tags (Bonds, Forward, Funds, Future, MMIs, Option, Stocks, Swap, Equity Index, Credit Index, Securities restrictions, Parametric schedules, Debt pricing and yields, Credit Events, Stock Corporation, Central Securities Depository, Regulatory Agency), we mine their corresponding FIBO web pages for the following properties:
\begin{itemize}
    \item Definition
    \item Explanatory Note
    \item Generated Description
    \item Synonym(s)
\end{itemize}
We also iterate over their children (n+1) instances found under the "Direct subclasses" web page section and collect their associated definitions. We do the same for the grandchildren (n+2) of the predefined tags at which point we stop the recursion. From the collected definitions we create a corpus of definition/tag pairs whereby each definition is associated to its corresponding tag. Children and grandchildren definitions are associated to one of the parent tags we started with. We stop the recursion at (n+2) because iterating further causes an overlap between concept definitions related to one or more of the predefined tags resulting in imprecise tag associations depending on the order of the recursion. Limiting the recursion at the (n+2) stage effectively prevents noise addition caused by such overlaps.
The final compiled dictionary contains a total of 2015 definitions. These definitions are used to train domain-specific sentence representations.

\subsection{Phrase Representation Learning}

In this component, we combine two representational techniques: word embeddings and sentence embeddings. By concatenating both word and sentence vectors for a phrase (i.e. the group of words that make up a term), we hope to capture the syntactic and semantic properties of the financial domain while trying to reduce the ambiguity that comes with domain-specific representational learning. To achieve this practically, we pad the word embeddings vector of dimension 300 with zeroes to obtain a new word vector of size 768 identical to the size of the sentence embeddings vector without loss of information stored by the word vector. Then we concatenate both vectors by performing a term-by-term addition operation. Our final vector model is of size 768 and combines the information captured by both the sentence and word embeddings vectors.
\par For both models, The training is performed on an NVIDIA GeForce RTX 2070 with Max-Q Design
8GB GPU machine. The construction of each embeddings model is detailed in the following subsections.

\subsubsection{Sentence Representation Learning}

\FloatBarrier
\begin{figure}[!ht]
    \centering
    \noindent{\includegraphics{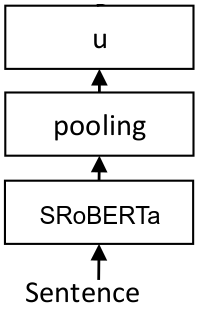}}
    \caption{Basic SentenceTransformer Architecture}
\end{figure}
\FloatBarrier

We use the version of SRoBERTa provided by Hugging Face\footnote{https://huggingface.co/sentence-transformers/nli-roberta-base-v2} and train it from scratch by adopting the method for training any BERT-like model on the STSBenchmark\footnote{http://ixa2.si.ehu.eus/stswiki/index.php/STSbenchmark} for the semantic similarity task\footnote{https://github.com/UKPLab/sentence-transformers/blob/master/examples/training/sts/training\_stsbenchmark.py}. We split our corpus into 70\% train set, 10\% dev set and 20\% test set. We also adopt the same model parameters as the STSBenchmark method for our training:
\begin{itemize}
    \item Training Batch Size: 8
    \item Number of Epochs: 4
\end{itemize}
\par To specialize our model, we use the extracted FIBO corpus of term definitions described earlier. In terms of pre-processing for each definition, we transform the text to lowercase and segment it into sentences based on newline and punctuation delimiters. The SentenceTransformer model has the following architecture depicted in Figure 2.
The depicted architecture consists of one RoBERTa layer and a pooling layer. We feed the input sentence or text into the RoBERTa transformer network. RoBERTa produces contextualized word embeddings for all input tokens in our text. Since we want a fixed-sized output representation (vector u), we need a pooling layer. Different pooling options are available, the most basic one being mean-pooling: we simply average all contextualized word embeddings RoBERTa produces. This gives us a fixed 768 dimensional output vector independently of how long our input text is.
\par For our training set, each definition is duplicated to match the total number of predefined tags we have. Each duplicate is then passed as an input sentence to the SentenceTransformer model along with a label indicating the semantic similarity with each of the tags. We use a label of 0.8 to indicate a positive example (i.e. corresponding to a matching definition-tag pair) and 0.3 for negative examples (i.e. all other duplicate instances of the definition with the remaining mismatched tags). The labels are chosen to be sufficiently far apart in value to discriminate especially for ambiguous terms. We feed the model a total of 317101 definition-label pairs.

\subsubsection{Word Representation Learning}

We augment the corpus compiled from the FIBO web pages with the English prospectuses set and use a FastText \cite{bojanowski2017enriching} model to generate custom domain-specific word embeddings. Between the two versions of custom word2vec models with dimensions 100 and 300 provided by the shared task organizers, the model with dimension 300 outperforms the smaller model. We use this as our starting point to generate two custom embeddings models, the first based on word2vec and the second on FastText, both of dimension 300, using our extracted corpus and compare their performance in the classification task. The results are detailed in Section 3.

\subsection{Classification Methods}

Sentence and word representations are used as features to train classifiers for term classification. A term is represented by the sum of phrase (sentence + word) embeddings for each word contained in the term. To find the best classifiers, we test two widely used classification methods: \textit{Logistic Regression} and \textit{Random Forest}.
Experimental studies will be discussed in Section 3.

\section{Experiments}

In this section we describe the data provided by the shared task organizers. We then provide details on the empirical experiments we performed and present our final results.

\subsection{Data Description}

The training data provided by the task organizers contains a total of 1050 entries where each entry consists of a term and its corresponding label. A label can be one of the 17 predefined tags. For the test data, there are 326 entries of terms to be correctly classified into the correct tag. The main difficulty in this classification task lies in the tag distribution: the chosen labels are not at the same ontological level as Figure 3 demonstrates.

\FloatBarrier
\begin{figure}[!ht]
    \centering
    \noindent{\includegraphics[width=0.5\textwidth, height=0.5\pdfpageheight, keepaspectratio]{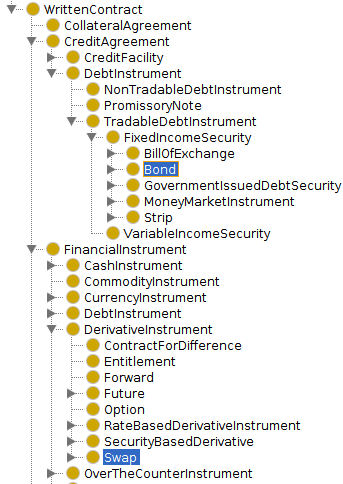}}
    \caption{An Example of Label Ontology Hierarchy}
\end{figure}
\FloatBarrier

The hierarchy shows that while some labels like Forward, Future, Option and Swap are on the same level, they are not aligned with other labels like Bond. The same case can be made for Central Securities Depository and Regulatory Agency in the FIBO ontology. The issue indicates that labels cannot be learned from a simple \textit{IS-A} relationship.
\par To tackle this problem, we enlarge the scope of our mining while collecting data from the FIBO web pages. Instead of limiting our collections to direct subclasses of the predefined tags, we search for "Instances" under "Ontological characteristic" which allows us to enrich our corpus both vertically and horizontally and expand term relations as much as possible by capturing their semantic connections.  

\subsection{Results and Analysis}

We design our experiments in order to determine the best model for each of the components in our proposed framework approach.
\par For the sentence embeddings module, we pit our SRoBERTa model against other well-performing models that we pre-train using the same setup described in Section 2.2. The selection of models is done based on our computational limitations as well as the grid\footnote{https://www.sbert.net/docs/pretrained\_models.html} proposed in the official SentenceTransformers Documentation. We select paraphrase-mpnet-base-v2, paraphrase-MiniLM-L6-v2 and paraphrase-distilroberta-base-v2 as the main competitors to SRoBERTa. To measure the performance of each custom trained model, we treat the classification problem like a semantic similarity task and use cosine similarity to find the best label for each term embedding. We evaluate model performance based on the metrics proposed by the shared task organizers. The results are shown in Table 1.

\begin{table}[ht]
\begin{center}
\begin{tabular}{|l|r|r|}
\hline \bf Model & \bf Accuracy & \bf Mean Rank \\ \hline
paraphrase-mpnet-base-v2 & 0.68 & 1.93 \\
paraphrase-MiniLM-L6-v2 & 0.65 & 2.44 \\
paraphrase-distilroberta-base-v2 & 0.63 & 2.65 \\
\textbf{SRoBERTa} & \textbf{0.73} & \textbf{1.61} \\
\hline
\end{tabular}
\end{center}
\caption{Sentence Embeddings Evaluation}
\end{table}

The empirical experiment is consistent with the choice of model since SRoBERTa is specialized in tasks like clustering or semantic search.
\par For the word embeddings model selection, we adopt the same baseline component proposed by the task organizers and composed of an embeddings module used as a feature vector and a logistic regression term classifier. The classifier is fixed in this experiment. The type of word embeddings is the only variable. The word2vec-100 and word2vec-300 models are the ones proposed by the task organizers and trained on the English prospectuses set (which we'll call Base). The c-word2vec-300 and c-fasttext-300 models are the models trained on our custom corpus (which we'll call Custom) comprised of the FIBO term definitions and the English prospectuses. Note that each result is the average of 5 runs and the train/test ratio is 80\%/20\%. The results are presented in Table 2.

\begin{table}[ht]
\begin{center}
\begin{tabular}{|l|r|r|r|}
\hline \bf Model & \bf Corpus & \bf Accuracy & \bf Mean Rank \\ \hline
word2vec-100 & Base & 0.76 & 1.43 \\
word2vec-300 & Base & 0.77 & 1.41 \\
c-word2vec-300 & Custom & 0.78 & 1.40 \\
\textbf{c-fasttext-300} & \textbf{Custom} & \textbf{0.82} & \textbf{1.33} \\
\hline
\end{tabular}
\end{center}
\caption{Word Embeddings Evaluation}
\end{table}

While training on the custom corpus enhances model performance, the results validate our choice of FastText as it outperforms word2vec due to the model's capability to retain subword information which results in better learning.
\par Finally, in order to improve term classification, we empirically study the performance of the classifier component by reversing the conditions of our previous experiment: we fix the feature vector to our best word embeddings model (c-fasttext-300) and vary the classifier. We keep the train/test split at 80\%/20\% and perform 5 runs. The results are displayed in Table 3.

\begin{table}[ht]
\begin{center}
\begin{tabular}{|l|r|r|}
\hline \bf Model & \bf Accuracy & \bf Mean Rank \\ \hline
Random Forest & 0.80 & 1.45 \\
\textbf{Logistic Regression} & \textbf{0.82} & \textbf{1.33} \\
\hline
\end{tabular}
\end{center}
\caption{Classifier Evaluation}
\end{table}

From this experimental study, we find that complicated classifiers like Random Forest achieve worse
performances than linear classifiers, so we select Logistic Regression as the classifier in our submitted systems. This observation shows that models that learn linear boundaries tend to perform better for this type of task.

\subsection{System Submissions}

In our submitted systems, we use the SRoBERTa model trained on the extracted corpus from the FIBO web pages. We submit a first system composed only of SRoBERTa and a classifier to study the performance of specialized sentence representations on this type of task. We use the constructed vector resulting from the concatenation of both the sentence model and the c-fasttext-300 word model as feature vector to the classifier in our second submission to study the effect of combining sentence and word information in what we refer to as phrase representation learning. Logistic regression is used in both submissions as the classifier. The final results are reported in Table 4. ACC is short for Accuracy and MR is short for Mean Rank.
\par \textit{yseop\_1} In this submission, we combine SRoBERTa as a feature vector with a Logistic Regression classifier. The dimension of representation for each term is 768.
\par \textit{yseop\_2} In this submission, we concatenate the SRoBERTa model with the padded c-fasttext-300 to produce a feature vector of size 768. We feed the resulting feature vector to a Logistic Regression classifier.

\begin{table}[ht]
\begin{center}
\begin{tabular}{|l|l|l|l|l|}
\hline
 & \multicolumn{2}{l|}{\bf Train Data} & \multicolumn{2}{l|}{\bf Test Data} \\ \hline
\bf System & \bf ACC & \bf MR & \bf ACC & \bf MR \\ \hline
yseop\_1 & 0.871 & 1.275 & 0.883 & 1.236 \\ \hline
\textbf{yseop\_2} & \textbf{0.883} & \textbf{1.234} & \textbf{0.917} & \textbf{1.141} \\ \hline
\end{tabular}
\end{center}
\caption{Final System Submissions}
\end{table}

From our submissions, \textit{yseop\_2} performs best and ranks 2$^{nd}$ overall on both Average Accuracy and Mean Rank metrics in the shared task.

\subsection{Data Imbalance}

Another issue in this shared task is the problem of data distribution. By examining our system results, we observe that our framework performs consistently better for some labels than others. We investigate the reason for the poor performance on some labels by averaging the accuracy of matched labels, i.e. labels that were correctly classified as the best choice for an entry, over 5 runs for our best system. The analysis yields:
\begin{itemize}
    \item Central Securities Depository: 78.45\%
    \item Credit Index: 79.28\%
    \item Bonds: 87.80\%
    \item Credit Events: 60\%
    \item Funds: 84.31\%
    \item Stock Corporation: 66.80\%
    \item Regulatory Agency: 90.76\%
    \item Debt pricing and yields: 97.94\%
    \item Equity Index: 96.25\%
    \item Swap: 75.18\%
    \item Option: 100\%
    \item Stocks: 45.71\%
    \item Future: 88.46\%
    \item Securities restrictions: 80\%
    \item Parametric schedules: 82.46\%
    \item MMIs: 18.30\%
    \item Forward: 57.38\%
\end{itemize}

This preliminary analysis reveals that while some labels like Option and Equity Index are over-expressed (the model predicts them correctly most of the times), others such as MMIs and Stocks are severely under-expressed. These results lead us to examining the composition of our train set distribution:
\begin{itemize}
    \item Central Securities Depository: 10.19\%
    \item Credit Index: 12.29\%
    \item Bonds: 5.24\%
    \item Credit Events: 1.71\%
    \item Funds: 2.1\%
    \item Stock Corporation: 2.38\%
    \item Regulatory Agency: 19.52\%
    \item Debt pricing and yields: 5.52\%
    \item Equity Index: 27.24\%
    \item Swap: 3.43\%
    \item Option: 2.29\%
    \item Stocks: 1.62\%
    \item Future: 1.81\%
    \item Securities restrictions: 0.75\%
    \item Parametric schedules: 1.43\%
    \item MMIs: 1.62\%
    \item Forward: 0.86\%
\end{itemize}

The distribution shows a discrepancy in label expression that may explain the over-prediction of certain labels whenever the model makes a wrong prediction. However, the train data is one source of our learning and is active at the classification component of our framework. The other main source of representation is the extracted corpus collected from FIBO. We propose to analyze the distribution of label occurrences in the corpus based on the definitions collected:
\begin{itemize}
    \item Central Securities Depository: 0.45\%
    \item Credit Index: 0.20\%
    \item Bonds: 13.30\%
    \item Credit Events: 4.27\%
    \item Funds: 4.81\%
    \item Stock Corporation: 2.18\%
    \item Regulatory Agency: 3.24\%
    \item Debt pricing and yields: 10.72\%
    \item Equity Index: 0.35\%
    \item Swap: 4.57\%
    \item Option: 6.00\%
    \item Stocks: 38.41\%
    \item Future: 2.43\%
    \item Securities restrictions: 3.57\%
    \item Parametric schedules: 3.82\%
    \item MMIs: 0.74\%
    \item Forward: 0.94\%
\end{itemize}

The last set of results shows the number of times a label is expressed is not sufficient to guarantee a good model performance. Some labels that are well expressed are under-represented in terms of definitions with respect to others. This effectively splits the data distribution problem in two ways: the first is balancing uner-represented labels using techniques such as SMOTE and the second enriching definitions for some labels to improve predictions using external sources. Both methods merit further exploration.

\section{Conclusion}

In this paper, we studied the task of hypernym identification in the financial domain. We trained a a phrase representation learning model by specializing and combining a SRoBERTa sentence embeddings model and a FastText word embeddings model on a relatively small data set. We also enriched the provided data by collecting term definitions and term relations for the proposed hypernyms. Our approach shows that it is possible to specialize a domain-specific model by combining sentence and word models with a linear classifier on a relatively small corpus. It would be interesting to explore future possibilities by exploiting other domain resources or enhancing under-represented labels and studying their impact on domain-specific learning.

\bibliographystyle{named}
\bibliography{ijcai21}

\end{document}